\documentclass[letterpaper, 10 pt, conference]{ieeeconf}
\IEEEoverridecommandlockouts
\let\labelindent\relax
\usepackage{enumitem}
\usepackage{balance}
\usepackage[labelformat=simple]{subfig}
\usepackage[font={footnotesize}]{caption}
\usepackage{array}
\usepackage{textcomp}
\usepackage{mathtools, nccmath}
\usepackage{graphicx}
\usepackage{amsfonts}
\usepackage{amsmath}
\usepackage{autobreak}
\usepackage{amssymb}
\usepackage{algorithm}
\usepackage{algorithmic}
\usepackage{tikz}
\usepackage{arydshln}
\usepackage{multirow}
\usepackage{bm}
\usepackage{epstopdf}
\usepackage{cite}
\usepackage{siunitx}
\usepackage{xcolor}
\usepackage{float} 
\usepackage{stfloats}
\usepackage{lineno}
\usepackage{subfloat}
\usepackage{colortbl}
\usepackage{booktabs}
\usepackage{balance} 

\makeatletter
\let\NAT@parse\undefined
\makeatother
\usepackage[colorlinks=true,
            linkcolor=blue,
           ]{hyperref}

\allowdisplaybreaks[4]

\title{\LARGE \bf A Universal Vehicle-Trailer Navigation System with Neural Kinematics and Online Residual Learning}
\author{Yanbo Chen$^*$, Yunzhe Tan$^*$, Yaojia Wang, Zhengzhe Xu, Junbo Tan$^\dagger$, and Xueqian Wang$^\dagger$
\thanks{$^*$ indicates equal contribution.}
\thanks{$^\dagger$ Corresponding authors:  Junbo Tan, Xueqian Wang.}
\thanks{This work was supported by the Natural Science Foundation of Shenzhen (No.JCYJ20230807111604008, No. JCYJ20240813112007010), the Natural Science Foundation of Guangdong Province (No.2024A1515010003), National Key Research and Development Program of China(No. 2022YFB4701400) and Cross-disciplinary Fund for Research and Innovation (No. JC2024002) of Tsinghua SIGS.}
\thanks{Yanbo Chen, Junbo Tan, and Xueqian Wang are with the Center for Artificial Intelligence and Robotics, Shenzhen International Graduate School, Tsinghua University, Shenzhen 518055, China, \tt \{cyb23@mails., tjblql@sz., wang.xq@sz.\}tsinghua.edu.cn}
\thanks{Yunzhe Tan, and Yaojia Wang are with the College of Artificial Intelligence at Harbin Institute of Technology, Shenzhen 518055, China, \tt\{210320330,220320604\}@stu.hit.edu.cn}
\thanks{Zhengzhe Xu is with the Department of Mechanical Engineering, University of Hong Kong, Pokfulam, Hong Kong, \tt{xuzz@connect.hku.hk}}
}

\begin{document}

\maketitle
\begin{abstract}

Autonomous navigation of vehicle-trailer systems is crucial in environments like airports, supermarkets, and concert venues, where various types of trailers are needed to navigate with different payloads and conditions.
However, accurately modeling such systems remains challenging, especially for trailers with castor wheels.
In this work, we propose a novel universal vehicle-trailer navigation system that integrates a hybrid nominal kinematic model—combining classical nonholonomic constraints for vehicles and neural network-based trailer kinematics—with a lightweight online residual learning module to correct real-time modeling discrepancies and disturbances.
Additionally, we develop a model predictive control framework with a weighted model combination strategy that improves long-horizon prediction accuracy and ensures safer motion planning.
Our approach is validated through extensive real-world experiments involving multiple trailer types and varying payload conditions, demonstrating robust performance without manual tuning or trailer-specific calibration.

\end{abstract}

\section{Introduction}
\label{sec:Introduction}

Autonomous navigation for vehicle-trailer systems has been extensively studied, primarily focusing on trailers equipped with rigid wheels (e.g., the flatbed and camper trailers) due to their well-understood kinematics described by nonholonomic constraints \cite{lundquist2006back,werling2013reversing,zhang2020universal}. 
However, in environments like airports, supermarkets, exhibition halls, and concert venues, trailers exhibit more diverse and complex motion characteristics. These trailers include platform trolleys with rigid wheels and transport devices like luggage trolleys, shopping carts, and flight cases featuring castor wheels. The kinematics of these trailers are influenced by unobservable wheel states, posing significant modeling and planning challenges. Previous studies typically address this by rigidly attaching trailers to the towing vehicle, neglecting castor influence \cite{xiao2022robotic,wang2021real}. However, rigid attachment results in excessive turning resistance for trailers with rigid wheels and significantly reduces maneuverability of large trailers in narrow spaces, limiting real-world applicability. Consequently, manual transportation remains prevalent in these scenarios. 

To address these limitations, we propose a universal vehicle-trailer navigation system designed for low-speed transportation tasks. Our system facilitates convenient changing of trailer types without manual parameter tuning, ensuring obstacle avoidance and safe navigation. We address the modeling complexities by integrating a neural network approach with nonholonomic kinematic constraints, supplemented by an online-learned residual kinematic network that adapts to various trailers, payload variations, and external disturbances. Based on this hybrid model, we introduce a Model Predictive Control (MPC) framework with a weighted combination strategy, enhancing long-horizon prediction accuracy and enabling efficient, secure motion planning.

\begin{figure}
    \centering
    \includegraphics[width=1.0\linewidth]{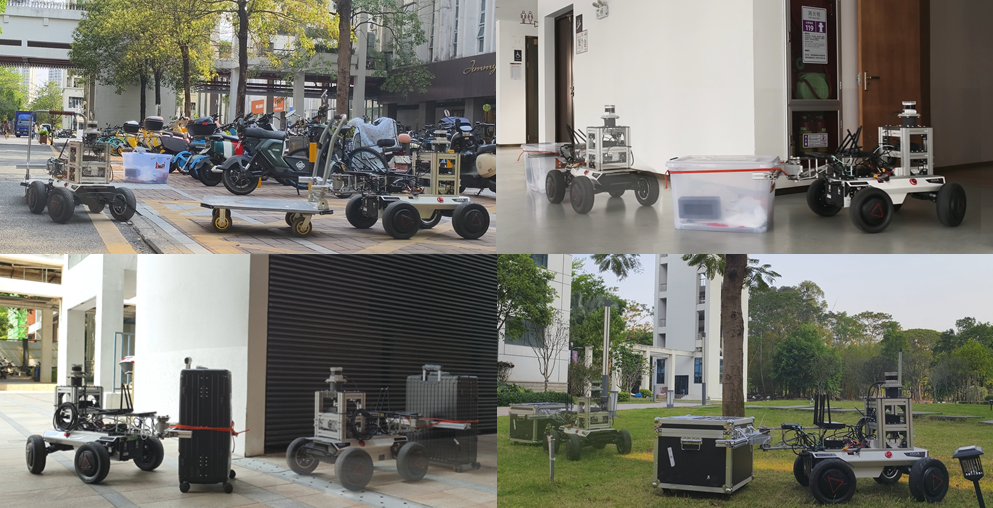}
    \caption{
    An Ackermann steering vehicle autonomously tows multiple types of trailers to their respective destinations, avoiding obstacles across diverse indoor and outdoor environments.
    }
    \label{fig:main}
\end{figure}

To sum up, this work offers the following contributions:
    \begin{itemize}
    \item We propose an efficient hybrid nominal kinematic model combining nonholonomic constraints for the vehicle and neural network-based kinematics for the trailer, and a lightweight online-trained residual kinematic network enhancing adaptability to modeling error and external disturbances.
    
    \item We develop an MPC framework utilizing a weighted model combination strategy to improve prediction accuracy over longer horizons, significantly enhancing trajectory tracking performance and system safety.
    
    \item We validate the proposed vehicle-trailer navigation system through extensive real-world experiments involving various trailer types, payload conditions, and complex environments. The results demonstrate the system's capability to autonomously adapt and robustly handle navigation tasks.
    \end{itemize}

\section{Related Work}
\label{sec:related}
The development of vehicle-trailer navigation systems has received significant attention due to their critical role in logistics and autonomous transportation. In the following, we briefly review relevant prior work categorized into three main areas: kinematic modeling, adaptive model error compensation, and motion planning.

\subsubsection{Kinematic Modeling}
\label{subsubsec:Kinematic Modeling}
Traditional trailer systems typically employ nonholonomic constraints to describe the kinematics of trailers with rigid wheels \cite{lundquist2006back,werling2013reversing,zhang2020universal}.
While effective for simple configurations, these models fail to generalize to trailers with castor wheels, where unobservable wheel states and dynamic interactions introduce nonlinearities.
Recent advancements in data-driven approaches, particularly neural network-based methods, have shown superior capability in modeling complex nonlinear dynamics and kinematics, as demonstrated in applications involving quadrotors \cite{bansal2016learning}, legged robot actuators \cite{hwangbo2019learning}, and vehicle models with tire friction \cite{spielberg2019neural}.
Inspired by these works, we propose a hybrid modeling approach that combines classical nonholonomic constraints for the vehicle with a neural network-based kinematic model for the trailer, achieving both computational efficiency and enhanced modeling accuracy.

\subsubsection{Adaptive Model Error Compensation}
\label{subsubsec:Adaptive Error Compensation}
Variations in trailer types, payloads, and external disturbances inevitably introduce errors into system models. Traditional methods, such as disturbance observers \cite{chen2015disturbance,lu2023robust} and gray-box system identification \cite{aastrom1971system,michalek2024scalable}, offer partial remedies but struggle to address real-time dynamic variations comprehensively \cite{korayem2021review}.
Recently, learning-based residual (error) models have emerged as powerful tools for handling modeling errors in real-time \cite{sun2021online,zeng2024adaptive,xue2024learning,chee2022knode,spielberg2021neural}. Motivated by these studies, our approach introduces a lightweight residual kinematics network trained online, effectively compensating for short-term modeling errors without imposing significant computational demands.

\subsubsection{Motion Planning}
\label{subsubsec:Motion Planning}
MPC is widely recognized as a robust framework for motion planning in autonomous systems due to its predictive capability and constraint management \cite{jian2023dynamic,zeng2021enhancing,chen2023quadruped}. It has also been extensively utilized in traditional trailer navigation \cite{ito2023configuration,bos2023mpc}. 
Recent studies have shown that integrating neural networks, including nominal and residual models, with MPC can significantly enhance prediction accuracy, tracking performance, and robustness in complex and uncertain environments \cite{salzmann2023real,ren2024npc,zeng2024adaptive,xue2024learning,chee2022knode}. Benefiting from our hybrid nominal model and lightweight residual network, we leverage MPC for effective long-horizon prediction and planning, thereby ensuring both safety and efficiency.

\section{Navigation System Overview}
\label{sec:overview}

\begin{figure}[t]
    \centering
    \includegraphics[width=8cm]{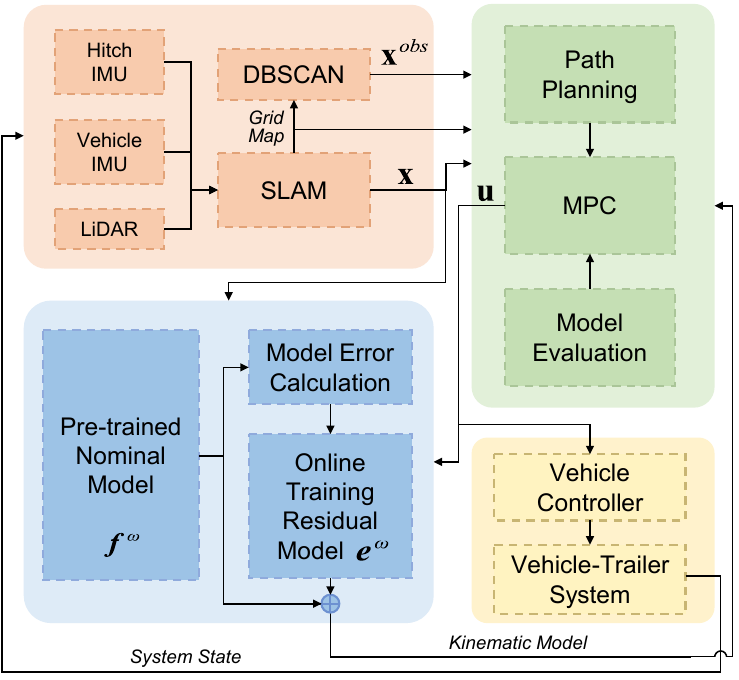}
    \caption{
    Overview of the navigation system. The framework integrates perception, learning-based kinematic model, planning, and control to enable safe and robust navigation.
    }
    \label{fig:system workflow}
\end{figure}

As shown in Fig.~\ref{fig:system workflow}, the proposed navigation system consists of four main parts: perception, learning-based kinematic model, planning, and control.

(a) \textbf{Perception}: The state of the towing vehicle is estimated using DLIO SLAM~\cite{chen2023direct}, which integrates LiDAR and IMU measurements. An additional IMU on the hitch provides additional measurements of the hitch angle, trailer yaw angle, and yaw rate. For obstacle avoidance, LiDAR point clouds are processed into a 2D occupancy grid map, and DBSCAN~\cite{ester1996density} is employed to cluster obstacles surrounding the system.
(b) \textbf{Learning-based Kinematic Model}: A nominal kinematic model of the vehicle-trailer system is pre-trained using pre-sampled trajectory data, as described in Sec. \ref{subsubsec:nominal model}. During navigation, discrepancies between the predicted and actual kinematics are computed in real-time, and a lightweight residual kinematic model is updated online using the most recent trajectory, as described in Sec. \ref{subsubsec:error model}. 
(c) \textbf{Planning}: Motion planning is implemented through an MPC framework, ensuring safe and kinematically feasible trajectories. An initial collision-free path for the Ackermann steering vehicle is generated using hybrid A*\cite{dolgov2010path}, providing a reference for MPC. The nominal and residual models jointly serve as MPC prediction components through a weighted combination strategy, with their relative predictive performances evaluated online. This online evaluation dynamically adjusts the weighting factors within MPC's constraints and cost functions to improve planning robustness, as detailed in Sec.~\ref{sec:Motion Planning}.
(d) \textbf{Control}: The planning framework employs receding horizon optimization, continuously updating the control commands sent to the vehicle, which uses a PID controller for closed-loop tracking.

\section{Modeling} \label{sec:Modeling}
\subsection{System State Representation}
\label{subsec:system state representation}

\begin{figure}[t]
    \centering
    \includegraphics[width=7cm]{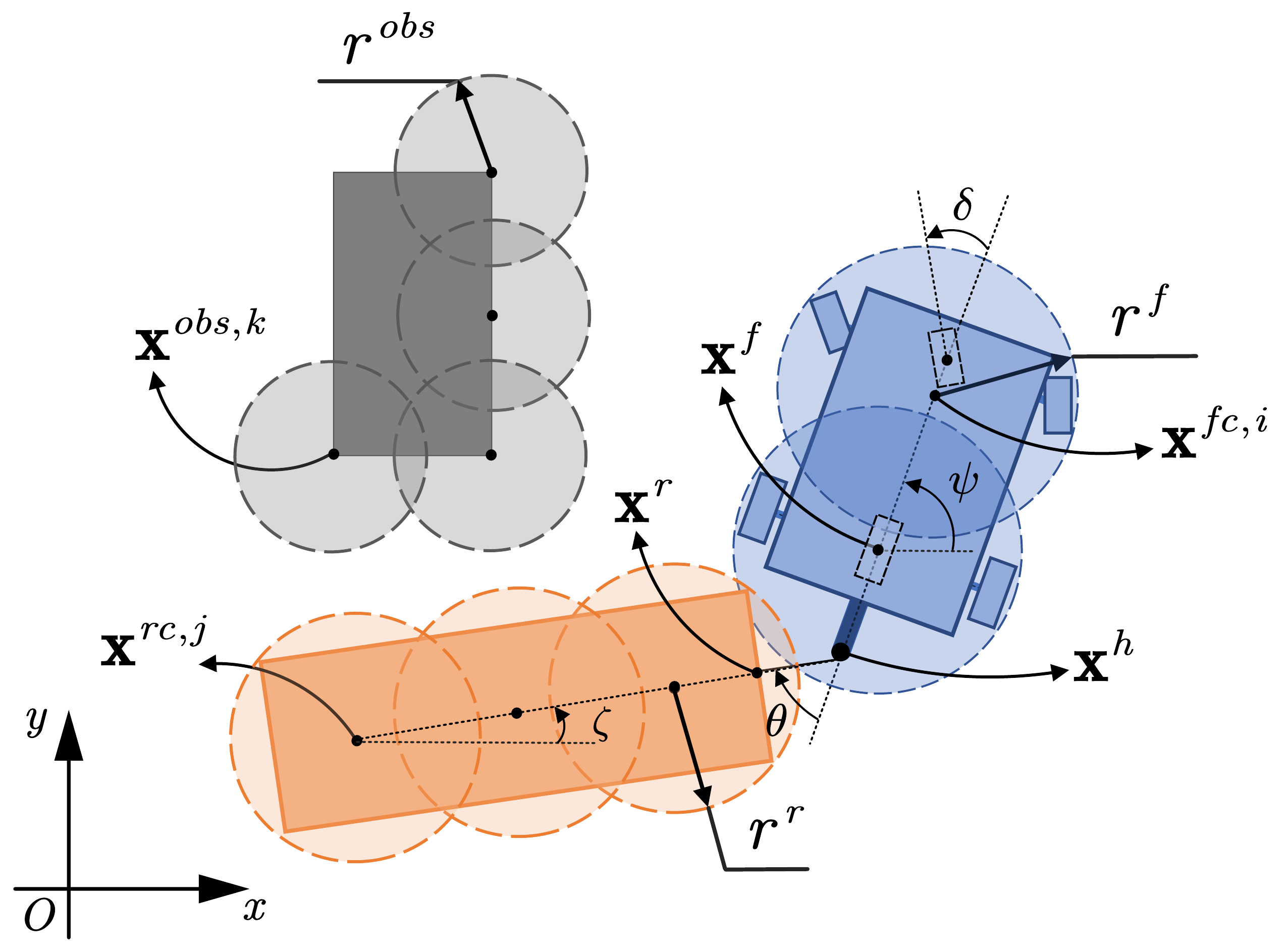}
    \caption{
    Definitions for the vehicle-trailer system. For obstacle avoidance, the vehicle, trailer, and obstacles are covered by multiple enclosing circles with radii $r^{f}$, $r^{r}$, and $r^{obs}$, respectively. Each circle has a center at positions $\mathbf{x}^{fc, i}$ ($i = 1, \dots, n^{fc}$) for the vehicle, $\mathbf{x}^{rc, j}$ ($j = 1, \dots, n^{rc}$) for the trailer, and $\mathbf{x}^{obs, k}$ ($k = 1, \dots, n^{obs}$) for obstacles.
    }
    \label{fig:system formulation}
\end{figure}

The vehicle-trailer system consists of an Ackermann steering vehicle towing a trailer through a hitch, as shown in Fig. \ref{fig:system formulation}. Since the system typically operates in relatively flat environments, it can be modeled in a 2D plane.

The state of the system is defined as:
\begin{equation}
    \mathbf{x} = [(\mathbf{x}^{f})^{\top},\, \psi,\, \zeta,\, {\omega}^{\zeta}]^{\top},
\end{equation}
where $\mathbf{x}^f = [x^f,\, y^f]^{\top}$ represents the position of the center of the vehicle's rear axle, $\psi$ and $\zeta$ denote the yaw angles for the vehicle and trailer respectively, and ${\omega}^{\zeta}$ is the trailer yaw rate.

Any other information about the system can be obtained from the state $\mathbf{x}$.
The hitch angle $\theta = \psi - \zeta$ is the relative yaw angle between the vehicle and the trailer. The hitch position can be got by $\mathbf{x}^{h} = \mathbf{x}^f - l_{fh}\mathbf{e}_{\psi}$, where $l_{fh}$ is the distance from the vehicle's rear axle to the hitch, and 
$\mathbf{e}_{(\cdot)} = [\cos{(\cdot)},\,\sin{(\cdot)}]^{\top},  \text{where } (\cdot) \text{ represents an arbitrary angle.}$
The trailer’s position is given by $\mathbf{x}^{r} = \mathbf{x}^{h} - l_{hr}\mathbf{e}_{\zeta}$, where $l_{hr}$ is the distance from the hitch to the trailer.

The system input is defined as:
\begin{equation}
    \mathbf{u} = [v,\, \delta]^{\top},
\end{equation}
where $v$ is the longitudinal velocity of the vehicle, and $\delta$ is the steering wheel angle. The vehicle yaw rate can be computed as:
$\omega^{\psi} = v\tan{(\delta)}/l$,
where $l$ is the wheelbase length of the vehicle.

For obstacle avoidance, the vehicle, trailer, and obstacles are covered by several circles, as shown in Fig.~\ref{fig:system formulation}. Each circle’s center positions and radii are defined appropriately to ensure safety margins during navigation.

\subsection{Hybrid Vehicle-Trailer Kinematics}
\label{subsec:vehicle-trailer kinematic model}

For a universal vehicle-trailer system, the behavior of different towed trailers depends on factors such as wheel types, payloads, and road conditions, making them difficult to model from first principles. 
To address this challenge, we propose a hybrid kinematic model consisting of two parts: (1) \textbf{nominal kinematics} $\boldsymbol{f}$, which combines the classical nonholonomic kinematic constraints of the towing vehicle with neural network-based kinematics of the trailers; and (2) \textbf{residual kinematics} $\boldsymbol{e}$, which compensates for discrepancies between the nominal and actual kinematics, including modeling errors $\varDelta \boldsymbol{f}$ and errors caused by external disturbances $\mathbf{d}$.
By integrating first-principles modeling and data-driven methods, this hybrid approach enhances both the accuracy and generalization capability of the model. 
It is formally formulated as:
\begin{equation}
    \mathbf{x}_{k+1} = \boldsymbol{f}(\mathbf{x}_{k-n_f:k},\mathbf{u}_{k-n_f:k}) + \boldsymbol{e}(\mathbf{x}_{k-n_f:k},\mathbf{u}_{k-n_f:k}).
    \label{equ:hybrid_model}
\end{equation}

We use the notation
$
\mathbf{z}_{a:b}=\left[ \begin{matrix}
	\mathbf{z}_{a}^{\top}&		\mathbf{z}_{a+1}^{\top}&		\cdots&		\mathbf{z}_{b}^{\top}\\
\end{matrix} \right] ^{\top}
$
to denote a stacked vector of elements spanning from index \(a\) to \(b\). 
Accordingly, in (\ref{equ:hybrid_model}), $\mathbf{x}_{k-n_f:k}$ and $\mathbf{u}_{k-n_f+1:k}$ represent the past $n_f+1$ frames of historical system states and control inputs, respectively.
Introducing historical information allows the model to implicitly encode unmeasured dynamic states (e.g., velocity and acceleration) and unknown parameters (e.g., trailer type and payload).

\subsubsection{Nominal Kinematics}
\label{subsubsec:nominal model}

Directly using a large-scale neural network to model the full kinematics of the vehicle-trailer system would require extensive training data and computational resources, and would significantly increase the computational complexity of optimization-based planning, which may be unaffordable for onboard computers with limited computing power. 
For the towing vehicle, the classical nonholonomic Ackermann steering model provides accurate and robust predictions of its kinematics in most conditions with minimal computational overhead. 
For the trailer, we only need to predict the yaw rate $\omega^{\zeta}$, which can be achieved with a relatively small neural network.
Based on these considerations, the nominal kinematic model is formulated as:

\begin{equation}
\boldsymbol{f}\left(
\begin{aligned}
&\mathbf{x}_{k-n_f:k}, \\
&\mathbf{u}_{k-n_f:k}
\end{aligned}
\right)
\!=\!
\begin{bmatrix}
x_k^f +  v_k \cos(\psi_k)\Delta t \\[4pt]
y_k^f +  v_k \sin(\psi_k)\Delta t \\[4pt]
\psi_k + \dfrac{v_k \tan(\delta_k)\Delta t}{l} \\[4pt]
\zeta_k + \omega_k^{\zeta}\Delta t \\[4pt]
\boldsymbol{f}^{\omega}\left(
  \theta_{k-n_f:k},\,
  \omega^{\zeta}_{k-n_f:k},\,
  \mathbf{u}_{k-n_f:k}
\right)
\end{bmatrix},
\label{equ:nominal_model}
\end{equation}
where $\Delta t$ denotes the time step, and $\boldsymbol{f}^{\omega}$ is a Multi-Layer Perceptron (MLP) neural network that predicts $\omega^{\zeta}$ based on the historical hitch angles $\theta_{k-n_f:k}$, trailer yaw rates $\omega^\zeta_{k-n_f:k}$, and control inputs $\mathbf{u}_{k-n_f:k}$.
This design choice is made because the trailer yaw rate is only affected by the relative configuration between the trailer and the towing vehicle, as well as the applied control inputs.

\subsubsection{Residual Kinematics}
\label{subsubsec:error model}

Although the nominal kinematics achieve good prediction performance under many conditions, unseen trailer types, varying payloads, and external disturbances can still introduce significant modeling errors. 
In particular, cumulative errors over long-term predictions can further degrade the accuracy (see Experiment~\ref{subsec:model evaluation}).

Instead of scaling up the nominal kinematics network, which would increase computational complexity, we introduce a residual kinematic model inspired by disturbance observer designs in control theory. 
It focuses on errors related to the trailer kinematics while ignoring the towing vehicle errors due to the high accuracy of the nonholonomic model, and is formulated as:

\begin{equation}
\boldsymbol{e}\left(
\begin{aligned}
&\mathbf{x}_{k-n_f:k}, \\
&\mathbf{u}_{k-n_f:k}
\end{aligned}
\right)
\!=\!
\left[ \begin{array}{c}
	\mathbf{0}_4\\
	\boldsymbol{e}^{\omega}\left( \theta _{k-n_f:k},\,\omega _{k-n_f:k}^{\zeta},\,\mathbf{u}_{k-n_f:k} \right)\\
\end{array} \right] 
,
  \label{eq:error_model}
\end{equation}
where $\boldsymbol{e}^{\omega}$ is a small-scale MLP network that dynamically compensates for modeling discrepancies of $\omega^\zeta$ through online learning. 

\subsubsection{Neural Networks Training Strategy}
\label{subsubsec:model_training}

The performance of the MPC-based planning module heavily relies on the accuracy of the kinematic model, particularly over long prediction horizons. 
For the neural network component of the system kinematics, it is crucial to ensure accurate predictions not only for the next single-step state but also across the entire $N$-step prediction horizon. 
To mitigate error accumulation during long-term rollouts, we adopt a rolling prediction strategy during training, as illustrated in Algorithm \ref{alg:Train_one_epoch}.
Specifically, for each batch of data, we perform $N$-step forward propagation through the network for all samples. The Mean Squared Error (MSE) is computed at each prediction step and accumulated over the entire horizon to form the training loss. 

\begin{figure}[t]
    \centering
    \includegraphics[width=1\linewidth]{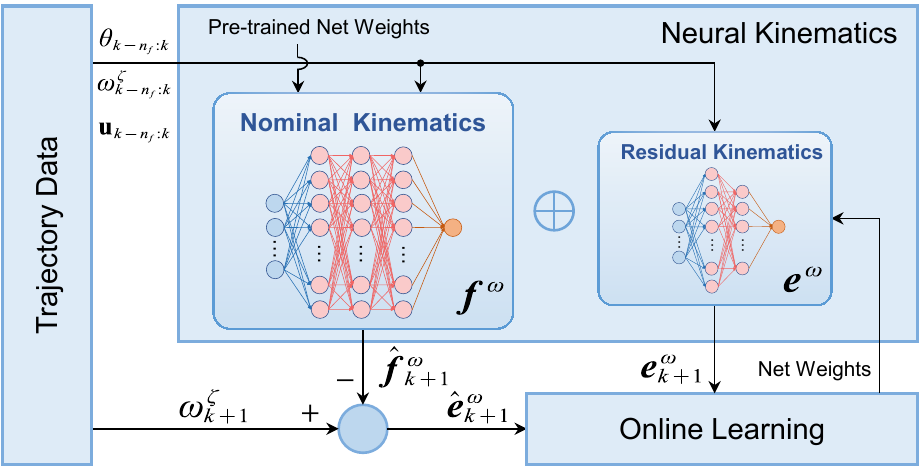}
    \caption{
    Structure of the proposed neural network kinematics for the trailer, where the nominal kinematics weights are pre-trained offline and the residual kinematics weights are updated online using recent trajectory.
    }
    \label{fig:online_training}
    \vspace{-0.3cm}
\end{figure}

\begin{algorithm}
    \caption{Rolling Prediction Training} 
    \label{alg:Train_one_epoch}  
    \begin{algorithmic}[1]
        \REQUIRE Batches $\mathcal{B}=\{\mathbf{b}_1, \mathbf{b}_2, \cdots, \mathbf{b}_m\}$, where $\mathbf{b}_i = (\mathbf{x}_{-n_f:N}^i, \mathbf{u}_{-n_f:N}^i)$ 
        \STATE $loss \gets 0$
        \STATE $trainer$.$\text{zero\_grad}()$  
        \FOR{each $i \in [1, m]$}  
            \STATE $\mathbf{x}_{-n_f:N}, \mathbf{u}_{-n_f:N} \gets \mathbf{b}_i$
            \STATE $\mathbf{\hat{x}}_{-n_f:0} \gets \mathbf{x}_{-n_f:0}$
            \FOR{each $k \in [0, N)$}
                \STATE $\mathbf{\hat{x}}_{k+1} \gets \boldsymbol{f}(\mathbf{\hat{x}}_{k-n_f:k}, \mathbf{u}_{k-n_f:k})$  
                \STATE $loss \gets loss + \mathrm{MSE}(\hat{\boldsymbol{f}}^{\omega}_{k+1}, {\omega}^{\zeta}_{k+1})$ 
            \ENDFOR
        \ENDFOR
        \STATE $loss$.$\text{backward}()$  
        \STATE $trainer$.$\text{step}()$
    \end{algorithmic}
\end{algorithm}

As illustrated in Fig. \ref{fig:online_training}, the weights of $\boldsymbol{e}^{\omega}$ are updated online using a sliding window of recent trajectory data. 
Specifically, at each time step $k$, the model is trained using the state sequence $\mathbf{x}_{k-n_t:k}$ and control sequence $\mathbf{u}_{t-n_t:t}$. 
By continuously updating online, the residual kinematics can adapt to short-term variations and external disturbances, providing a lightweight yet effective correction to the nominal kinematics without sacrificing computational efficiency.

\section{Motion Planning} \label{sec:Motion Planning}
\subsection{Problem Formulation}\label{subsec:Problem Formulation}
Motion planning is essential for ensuring the safe and efficient navigation of the vehicle-trailer system. We utilize MPC with an $N$-step horizon to optimize the control inputs. Specifically, we formulate the optimization problem with a multi-objective cost function ((\ref{equ:cost}), see Sec. \ref{subsec:Cost Function})) and constraints ((\ref{equ:init_x_cons})-(\ref{equ:input_bound_cons}), see Sec. \ref{subsec:Constraints}) as follows:

\begin{subequations}
\label{equ:MPC}
\begin{align}
    \underset{\tilde{\mathbf{x}}, \tilde{\mathbf{u}}}{\min}\quad &l_t\left( \tilde{\mathbf{x}}_N \right) +\sum_{k=1}^{N-1}{\left( l_s\left( \tilde{\mathbf{x}}_k \right) +l_i\left( \tilde{\mathbf{u}}_{k:k+1} \right) +l_o\left( \tilde{\mathbf{x}}_k \right) \right)} 
    \label{equ:cost}\\[6pt]
    \text{s.t.}\quad 
    &\tilde{\mathbf{x}}_{-n_f:0}=\mathbf{x}_{t-n_f:t},
    \label{equ:init_x_cons}\\
    &\tilde{\mathbf{u}}_{-n_f:0}=\mathbf{u}_{t-n_f:t},
    \label{equ:init_u_cons}\\[6pt]
        &\begin{aligned}
            \tilde{\mathbf{x}}_{k+1} &= \boldsymbol{f}\left( \tilde{\mathbf{x}}_{k-n_f:k},\tilde{\mathbf{u}}_{k-n_f:k} \right) \\
            &+ \lambda_e\left( k \right) \boldsymbol{e}\left( \tilde{\mathbf{x}}_{k-n_f:k},\tilde{\mathbf{u}}_{k-n_f:k} \right), \\
            & \quad k=0,\cdots ,N-1
        \end{aligned}
    \label{equ:model_cons}\\[6pt]
    &d_{fr}(\tilde{\mathbf{x}}_k) \geqslant 0, \quad k=1,\dots,N
    \label{equ:obs_cons}\\[6pt]
    &\tilde{\mathbf{x}}_k\in \mathcal{X}, \quad k=1,\dots,N
    \label{equ:state_bound_cons}\\
    &\tilde{\mathbf{u}}_k\in \mathcal{U}, \quad k=1,\dots,N
    \label{equ:input_bound_cons}
\end{align}
\end{subequations}

\subsection{Constraints}\label{subsec:Constraints}
\subsubsection{Initial Constraints}
The trajectory considered in each optimization consists of a historical trajectory of length $n_f+1$ and a predicted future trajectory of length $N$. (\ref{equ:init_x_cons}) and (\ref{equ:init_u_cons}) ensure that the initial $n_f+1$ optimization elements match the measured historical trajectory at time $t$. 

\subsubsection{Model Constraints}
In (\ref{equ:model_cons}), the kinematic model described in Sec.~\ref{subsec:vehicle-trailer kinematic model} is used to predict future states of the system. Since the residual kinematics $\boldsymbol{e}$ primarily depends on short segments of recent trajectory data, it is more reliable over a relatively short prediction horizon near time $t$. 

To limit its influence, we introduce a weighting function $\lambda_e(k)$, defined as:
\begin{equation}
    \lambda_e(k) = s_e \cdot \max\!\left(1 - \frac{k}{n_c},\, 0\right),
    \label{equ:model_weight_func}
\end{equation}
where $n_c$ is the cutoff step, $\lambda_e(k)$ decays linearly to zero at $k = n_c$, and $s_e \in \{0,1\}$ is a binary switch variable indicating whether the residual kinematic model should be activated. 

The value of $s_e$ is determined based on model performance over the past $n_e$ time steps. Specifically, let $\sigma_f$ and $\sigma_{fe}$ denote the MSE of ${\omega}^{\zeta}$ rolling predicted by the nominal model $\boldsymbol{f}$ and the unweighted combination $\boldsymbol{f} + \boldsymbol{e}$, respectively. If the ratio $\sigma_{fe} / \sigma_f$ is smaller than a predefined threshold $\epsilon$, then $s_e$ is set to 1; otherwise, it is set to 0. Note that during each optimization, $\lambda_e$ is a constant parameter, as $s_e$ is determined in advance.

\subsubsection{Obstacle Avoidance}
(\ref{equ:obs_cons}) enforces a non-negative minimum distance between the system and obstacles at each prediction step. Specifically, 
$d_{fr}\left( \tilde{\mathbf{x}}_k \right)$ is defined as:
\begin{align}
    d_{fr}\left( \tilde{\mathbf{x}}_k \right) =\min \bigg( 
    &\underset{i=1}{\overset{n^{fc}}{\min}}\,\,\underset{j=1}{\overset{n^{obs}}{\min}}\,\,d_{fc}\left( \tilde{\mathbf{x}}_{k}^{fc,i},\mathbf{x}^{obs,j}  \right) 
    \notag\\
    , 
    &\underset{i=1}{\overset{n^{rc}}{\min}}\,\,\underset{j=1}{\overset{n^{obs}}{\min}}\,\,d_{rc}\left( \tilde{\mathbf{x}}_{k}^{rc,i},\mathbf{x}^{obs,j} \right)  \bigg),  
    \label{equ:dfr}
\end{align}
where, $d_{fc}$ and $d_{rc}$ are given by:
\begin{align}
    d_{\left( fc|rc \right)}\left( \tilde{\mathbf{x}}_{k}^{\left( fc|rc \right),i},\mathbf{x}^{obs,j} \right) =\left\| \tilde{\mathbf{x}}_{k}^{\left( fc|rc \right) ,i}-\mathbf{x}^{obs,j}\right\|  
    \notag \\ -r^{\left( fc|rc \right)}-r^{obs}-d^{safe}.  
\end{align}

In this formulation, $\tilde{\mathbf{x}}_{k}^{\left( fc|rc \right) ,i}$ is derived from $\tilde{\mathbf{x}}_{k}$ denoting the centers of enclosing circles for predicted vehicle and trailer, $\mathbf{x}^{obs,j}$ are the centers of obstacle circles, which remains fixed during each optimization, $r^{\left( fc|rc \right)}$ and $r^{obs}$ are the radii of the corresponding circles as illustrated in Fig. \ref{fig:system formulation}, and $d^{safe}\geq 0$ represents the safety distance.

\subsubsection{State and Input Bounds}
(\ref{equ:state_bound_cons}) and (\ref{equ:input_bound_cons}) specify the constraints on the feasible sets for states (\(\mathcal{X}\)) and inputs (\(\mathcal{U}\)). In addition to the upper and lower bounds, \(\mathcal{X}\) also must satisfy \(\theta_l \leq \theta = \psi - \zeta \leq \theta_u\) to prevent self-collision between the trailer and vehicle. Meanwhile, \(\mathcal{U}\) must satisfy \(\omega^{\psi}_l \leq \omega^{\psi} = \frac{v \tan(\delta)}{l} \leq \omega^{\psi}_u\) to ensure that the vehicle's angular velocity remains within mechanical and safety limits.

\subsection{Cost Function}\label{subsec:Cost Function}

\subsubsection{Terminal Cost}
The terminal cost $l_t\left( \tilde{\mathbf{x}}_N \right) =\left\| \tilde{\mathbf{x}}_N-\mathbf{x}_{ter} \right\| _{\mathbf{Q}_{t}}^{2}$ is used to guide the system to the target state $\mathbf{x}_{ter}= \left[ x_{ter}, y_{ter}, \psi_{ter}, \zeta_{ter}, 0 \right]^{\top}$, where $\left\| \mathbf{x} \right\| _{\mathbf{Q}}^2 \coloneqq \mathbf{x}^{\top} \mathbf{Q} \mathbf{x}$, and the weight matrix $\mathbf{Q}$ is positive-definite.

\subsubsection{State Cost}

The state cost is defined as: 
$
    l_s\left( \tilde{\mathbf{x}}_k \right) =\left\| \tilde{\mathbf{x}}_{k}^{f}-\mathbf{x}_{k}^{f*} \right\| _{\mathbf{Q}_{ref}}^{2},
    \label{equ:state_cost}
$
where $\left\{\mathbf{x}_{k}^{f*}\right\}_{k=0}^{N}$ is the reference path from global path planning.

\subsubsection{Input Cost}
The input cost is defined as:
\begin{align}
l_i\bigl(\tilde{\mathbf{u}}_{k:k+1}\bigr)
= \|\tilde{\mathbf{u}}_k\|_{\mathbf{Q}_u}^2
+ \|\tilde{\mathbf{u}}_{k+1} - \tilde{\mathbf{u}}_k\|_{\mathbf{Q}_{du}}^2,
\end{align}
which penalizes both large control inputs and their rapid changes. The weight matrices $\mathbf{Q}_u$ and $\mathbf{Q}_{du}$ are diagonal, and their diagonal entries scale with the $\sigma_w$, which denotes the MSE of ${\omega}^{\zeta}$ rolling predicted by the weighted model (\ref{equ:model_cons}) over past $n_e$ steps. 

For trailers with castor wheels, high input accelerations can cause violent yaw rotations around the hitch, which significantly increase model uncertainty and may result in self-collision. Therefore, under conditions of high $\sigma_w$, enforcing stronger penalties on input rate changes through $\mathbf{Q}_{du}$ becomes particularly important to maintain system safety.

\subsubsection{Obstacle Distance Cost}
We introduce an obstacle avoidance cost to keep the vehicle and trailer away from obstacles:
\begin{align}
l_o\left( \tilde{\mathbf{x}}_k \right) &=\sum_{i=1}^{n^{fc}}{\sum_{j=1}^{n^{obs}}{\lambda _f\sigma _w\exp \left( -d_{fc}\left( \tilde{\mathbf{x}}_{k}^{fc,i},\mathbf{x}^{obs,j} \right) /\gamma _f \right)}}
\notag \\
&+\sum_{i=1}^{n^{rc}}{\sum_{j=1}^{n^{obs}}{\lambda _r\sigma _w\exp \left( -d_{rc}\left( \tilde{\mathbf{x}}_{k}^{rc,i},\mathbf{x}^{obs,j} \right) /\gamma _r \right)}}
\end{align}
where $\lambda _f$ and $\lambda _r$ are weighting parameters, and $\gamma_f$ and $\gamma_r$ control the decay rate of the function. This cost ensures that the system maintains a greater distance from obstacles when the model prediction error is large.
\section{Experiments}

\subsection{Experimental Setup}
\begin{figure}[!tbp]
    \centering
    \includegraphics[width=1\linewidth]{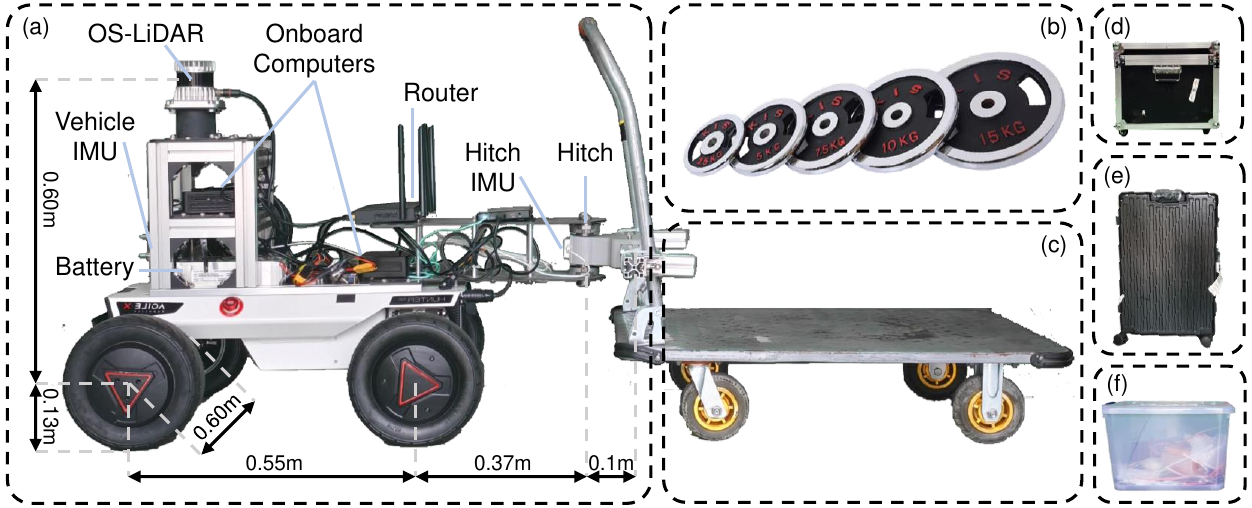}
    \caption{Experimental robot platform. The vehicle (HUNTER SE) is equipped with
    LiDAR, IMU, and two onboard computers. Various payloads (Fig.
    \ref{fig:hardware}(b)) and trailers (Fig. \ref{fig:hardware}(c)-(f)) are
    used for evaluation, including a platform trolley, a flight case, a suitcase,
    and a plastic wheelless storage box. }
    \label{fig:hardware}
\end{figure}

As shown in Fig. \ref{fig:hardware}(a), the experimental robot platform consists
of an Ackermann-steering mobile robot equipped with an OS0-128 LiDAR and a high-precision
9-axis IMU. An additional IMU is mounted at the hitch joint to measure the trailer’s
yaw angle and yaw rate. Two onboard computers (Intel NUC 11, i7-1165G7@2.80 GHz,
32 GB RAM) are used, one for perception and online residual kinematic model
training, and the other for planning and control.

To evaluate the system's performance under varying payloads and trailer types,
barbell plates of different weights are used, along with several types of trailers:
(i) a platform trolley ($0.9\,\mathrm{m}\times 0.6\,\mathrm{m}\times 0.2\,\mathrm{m}$)
with front castor wheels and rear rigid wheels is used to simulate the most
typical trailers (Fig. \ref{fig:hardware}(c)); (ii) a flight case ($0.5\,\mathrm{m}
\times 0.5\,\mathrm{m}\times 0.5\,\mathrm{m}$) and a suitcase ($0.45\,\mathrm{m}\times
0.28\,\mathrm{m}\times 0.73\,\mathrm{m}$), both equipped with four castor wheels,
representing fully maneuverable trailers (Fig. \ref{fig:hardware}(d-e)); (iii) a
plastic storage box ($0.65\,\mathrm{m}\times 0.50\,\mathrm{m}\times 0.42\,\mathrm{m}$)
without wheels is used to simulate towing of a non-rolling load (Fig.
\ref{fig:hardware}(f)).

All algorithms presented in this paper are integrated into the Robot Operating
System (ROS) and run on Ubuntu 20.04. The MPC problem is solved using the ACADOS
\cite{verschueren2021acados}. The prediction horizon is set to $N=30$, and the sampling
time is $\Delta t = 0.1\,\mathrm{s}$, which matches the update rate of the LiDAR
point cloud. The average computation time is approximately $20\,\mathrm{ms}$,
and the re-planning frequency is set to $10\,\mathrm{Hz}$.

\subsection{Data Collection and Training Details}
\label{subsec:data collection}
To train the nominal kinematic network of trailer, the vehicle towing either a
platform trolley or a flight case was remotely controlled under varying payloads,
surface conditions, speeds, and turning radii, and the system states and inputs
described in Sec. \ref{subsec:system state representation} were collected.
Historical information from the past 4 time steps ($n_{f}=3$) was concatenated
into the model input. In total, 40 trajectories were collected, each containing approximately
1500 time steps. The dataset was randomly split into training, validation, and
testing sets at a ratio of 8:1:1. The nominal kinematic network consists of three hidden
layers with dimensions of 64, 32, and 16, respectively. Tanh was adopted as the
activation function. The model was trained using PyTorch on an NVIDIA RTX 3090 GPU
with a batch size of 256. The Adam optimizer was employed with a learning rate decaying
from $10^{-2}$ to $10^{-5}$. The MSE of rolling prediction described in Sec. \ref{subsubsec:model_training}
was used as the loss function.

The residual kinematic network consists of two hidden layers with dimensions of
32 and 16, respectively, and is trained online using trajectory from the most recent
$n_{t}= 200$ time steps on the onboard computer's CPU. We use past $n_{e}= 15$
time steps data to evaluate the model's performance and set
$\epsilon=0.5 , n_{c}=15$ to calculate $s_{e}$ and $\lambda_{e}(k)$ in (\ref{equ:model_weight_func}).

\subsection{Neural Network Model Evaluation}
\label{subsec:model evaluation}

\subsubsection{Model Generalization Analysis}

\begin{figure}[!tbp]
    \centering
    \includegraphics[width=1\linewidth]{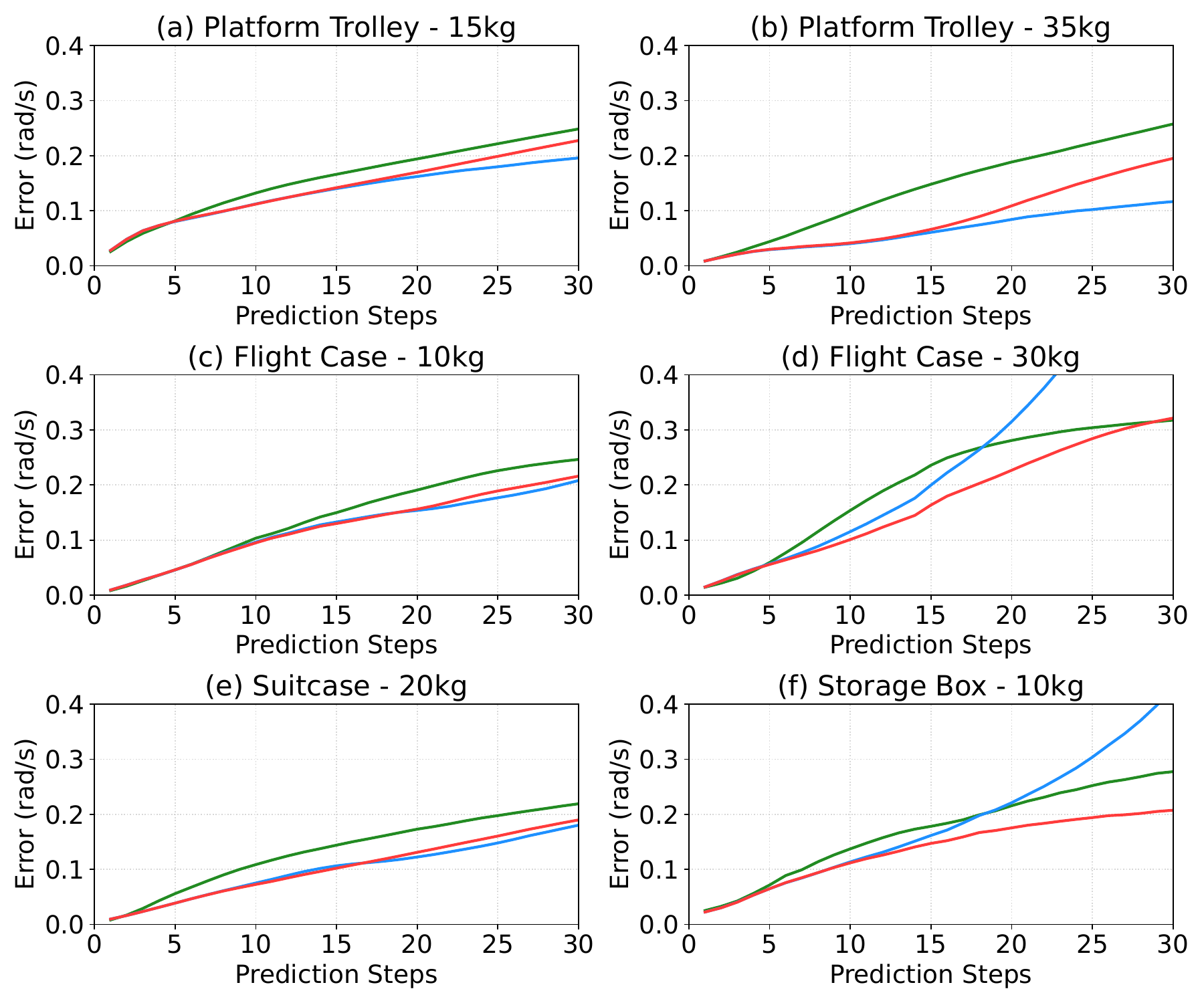}
    \includegraphics[width=1.0\linewidth]{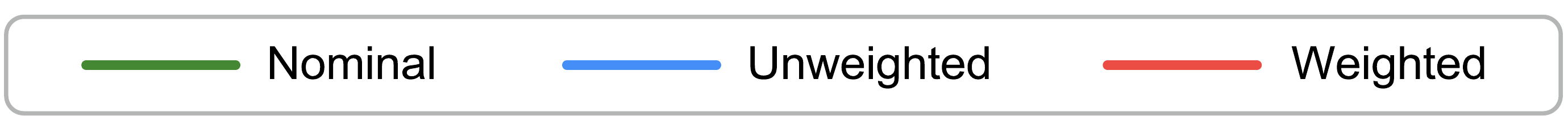}
    \caption{Multi-step rolling prediction RMSE of ${\omega}^{\zeta}$ using
    three models: the nominal model, the unweighted combination, and the weighted
    model. (a)-(b) Platform trolley with different payloads; (c)-(d) Flight case
    with different payloads; (e) Suitcase; (f) Wheelless storage box. }
    \label{fig:generalization}
\end{figure}

To evaluate the generalization capability of the neural network model and demonstrate
the effectiveness of the weighted model (\ref{equ:model_cons}), we conduct
experiments on a trajectory involving multiple 90-degree turns, where trailers are
prone to lateral slip, thus introducing significant model uncertainty. At each time
step, rolling predictions are performed for $1$ to $N$ steps using three
approaches: the nominal model $\boldsymbol{f}$, the unweighted combination
$\boldsymbol{f}+ \boldsymbol{e}$, and the weighted model. The Root Mean Squared
Error (RMSE) of ${\omega}^{\zeta}$ is calculated for each prediction step and
the results for six different trailer conditions are shown in Fig. \ref{fig:generalization}.

In conditions included in the training dataset, which are the platform trolley
and flight case with lower payloads (Fig.~\ref{fig:generalization}(a),(c)), the nominal
model exhibits good prediction performance, and the residual kinematic model
effectively compensates for small discrepancies. Due to the cutoff step being set
at $n_{c}=15$, the weighted model gradually becomes less accurate than the
unweighted combination after this step, but still outperforms the nominal model.

To evaluate generalization under payload variations, we increase the payloads of
the platform trolley and flight case beyond the maximum weight included in the
training set. The platform trolley, due to its rigid wheels, experiences a minimal
lateral slip and thus exhibits robustness to payload changes (Fig. \ref{fig:generalization}(b)).
For the flight case, the increased payload results in larger inertia, and the
castor wheels can not prevent lateral movement, leading to larger prediction errors
from the nominal model (Fig. \ref{fig:generalization}(d)). However, these errors
are effectively compensated by the residual kinematics. The unweighted
combination performs poorly in long-horizon predictions due to being trained on
a limited amount of historical data. This issue is effectively mitigated by the weighting
function.

The suitcase and box are used to evaluate the model's generalization to previously
untrained trailer types. The suitcase has similar kinematic characteristics to
the flight case, resulting in similar performance (Fig. \ref{fig:generalization}(c),
(e)). The box without wheels has different kinematic characteristics, leading to
a performance trend similar to that of the high-payload flight case (Fig.
\ref{fig:generalization}(d), (f)). Furthermore, the intersection between the unweighted
combination and the nominal model occurs after 15 steps, which justifies our
choice of $n_{c}=15$.

\subsubsection{Tracking Performance Analysis}

\begin{figure}[!t]
    \centering
    \begin{minipage}[t]{\linewidth}
        \centering
        \includegraphics[width=\linewidth]{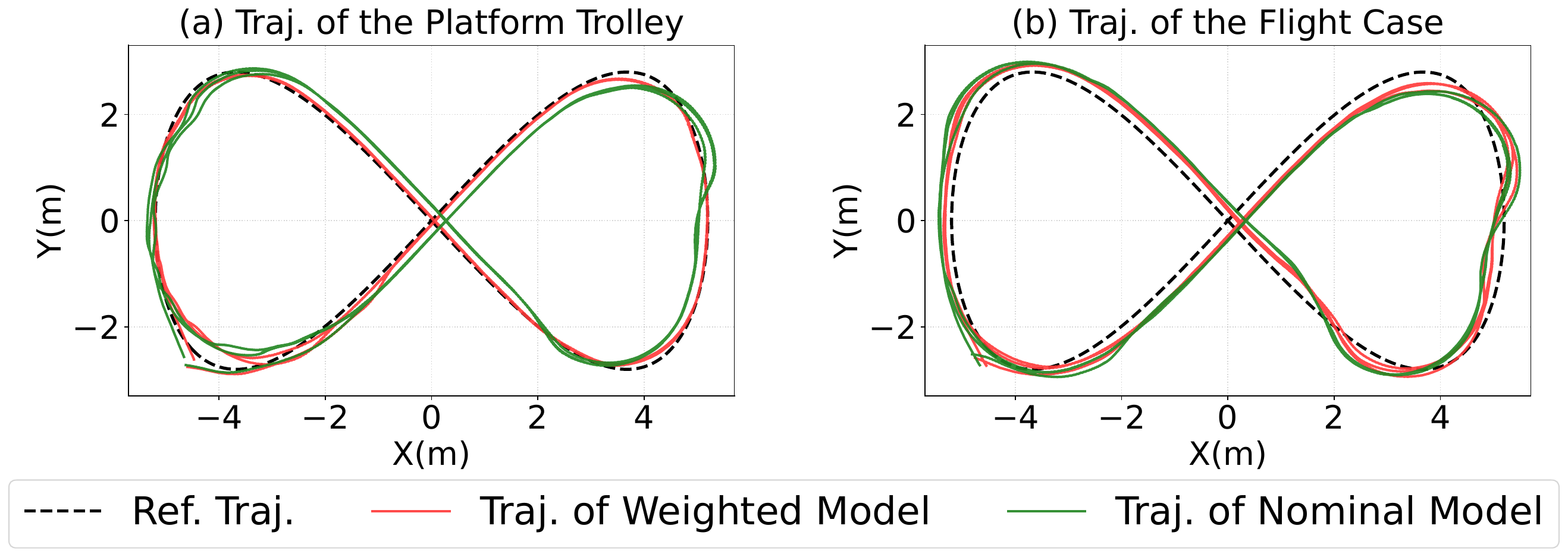}
        \vspace{-0.35cm}
    \end{minipage}
    \caption{Tracking trajectories of the trailers using the nominal model and
    weighted model within MPC.}

    \label{fig:tracking_trajectories}
\end{figure}

To evaluate the performance of the MPC with different models, we conduct trailer
trajectory tracking experiments. To eliminate the influence of other factors, we remove the obstacle avoidance constraint in (\ref{equ:MPC}) and replace the original cost function with a simplified objective:
$\left\| \tilde{\mathbf{x}}_{k}^{r}-\mathbf{x}_{k}^{r*}\right\|_{Q_{ref}}^{2}$, where
$\left\{\mathbf{x}_{k}^{r*}\right\}$ denotes the predefined 8-shaped trajectory.
We test the nominal model and weighted model as model constraints, using the platform
trolley and flight case as examples. The resulting tracking trajectories are
shown in Fig.~\ref{fig:tracking_trajectories}, and the tracking errors metrics
are presented in Table \ref{tab:tracking_error_stats}. The results indicate that
the MPC that uses the weighted model achieves better performance than the nominal
model.

\begin{table}[htbp]
    \centering
    \caption{Comparison of Tracking Errors Between Different Models}
    \begin{tabular}{@{}llccc@{}}
        \toprule \textbf{Trailer Type}             & \textbf{Model} & \textbf{Mean (m)} & \textbf{Std. Dev. (m)} & \textbf{Max (m)} \\
        \midrule \multirow{2}{*}{Platform Trolley} & Nominal        & 0.16              & 0.08                   & 0.36             \\
                                                   & Weighted       & \textbf{0.06}     & \textbf{0.07}          & \textbf{0.16}    \\
        \midrule \multirow{2}{*}{Flight Case}      & Nominal        & 0.22              & 0.07                   & 0.35             \\
                                                   & Weighted       & \textbf{0.16}     & \textbf{0.06}          & \textbf{0.25}    \\
        \bottomrule
    \end{tabular}
    \label{tab:tracking_error_stats}
\end{table}

\subsection{Autonomous Trailer Transportation Demonstration}
\label{subsec:navigation test}

\begin{figure*}[!t]
    \centering
    \includegraphics[width=1.0\linewidth]{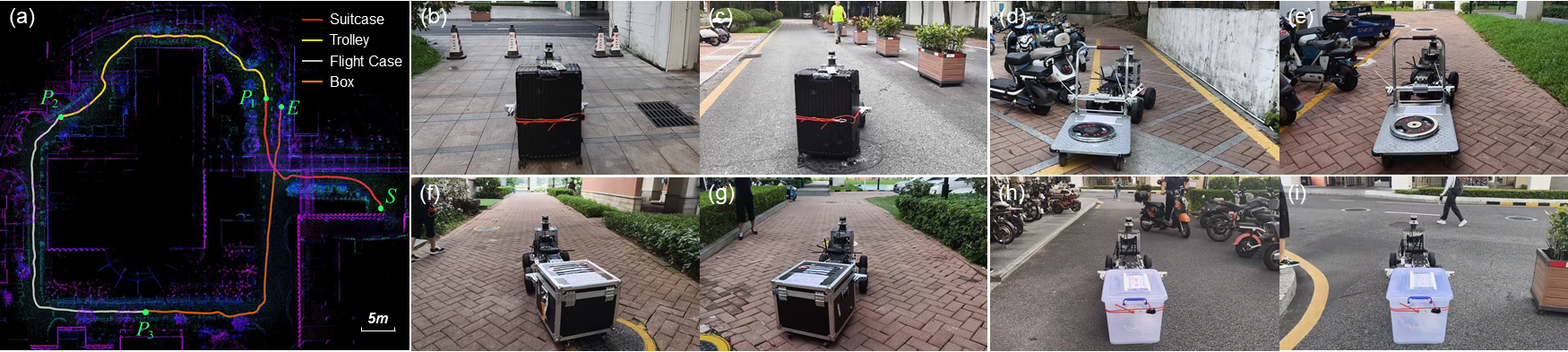}
    \caption{ Snapshots of the autonomous trailer transportation experiment demonstrating
    the proposed system's capability to navigate and deliver various types of trailers.
    (a) The point cloud map constructed via SLAM, with the trajectories of four trailer
    types, each covering approximately 50m. (b)-(c) The suitcase is towed through
    two 90-degree turns, across a road, and around obstacles such as traffic cones
    and planter boxes ($S \to P_{1}$). (d)-(e) The platform trolley successfully
    passes through a narrow corridor formed by a wall and parked bicycles ($P_{1}
    \to P_{2}$). (f)-(g) The flight case traverses a path with uneven brick pavement
    and manholes, introducing notable disturbances to the system ($P_{2}\to P_{3}$).
    (h)-(i) The box navigates around dynamic obstacles such as pedestrians and bicycles
    and reaches the final destination ($P_{3}\to E$). }
    \vspace{-0.4cm}
    \label{fig:real_world_snapshot}
\end{figure*}

To validate the robustness and generality of the proposed navigation system, we conducted
real-world trailer transportation experiments, as shown in Fig.~\ref{fig:real_world_snapshot}.
The system successfully navigates and delivers four different types of trailers
to their destinations, avoiding various static and dynamic obstacles without self-collisions.
Each trailer is handled without manual reconfiguration, demonstrating the system’s
ability to adapt to diverse trailer kinematics.

\begin{figure*}[t]
    \centering
    \subfloat[Suitcase.\label{fig:nn_1_rw}]{ \includegraphics[width=0.21\textwidth]{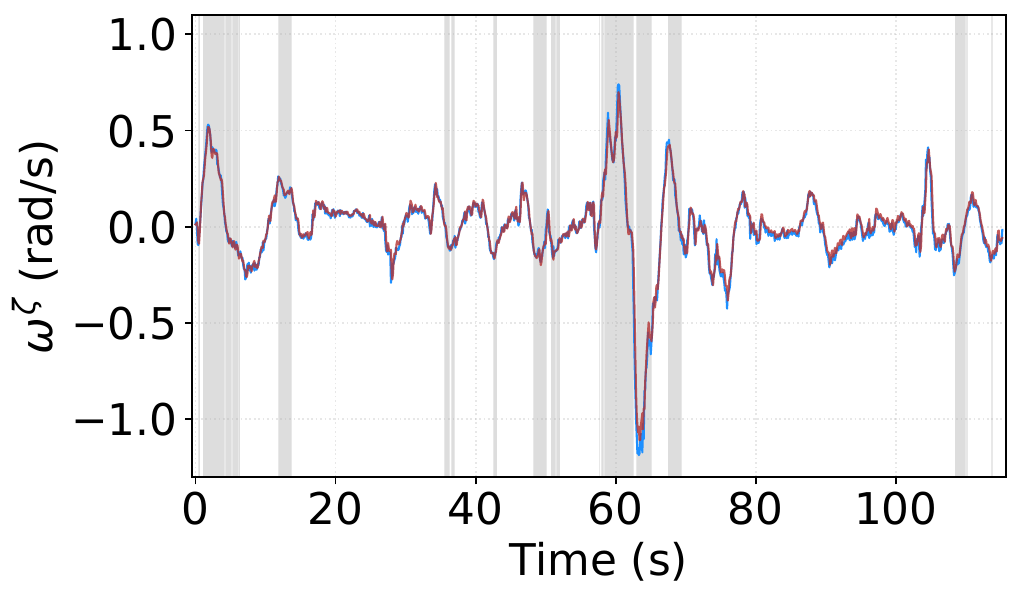} }
    \hfill \subfloat[Platform Trolley.\label{fig:nn_2_rw}]{ \includegraphics[width=0.21\textwidth]{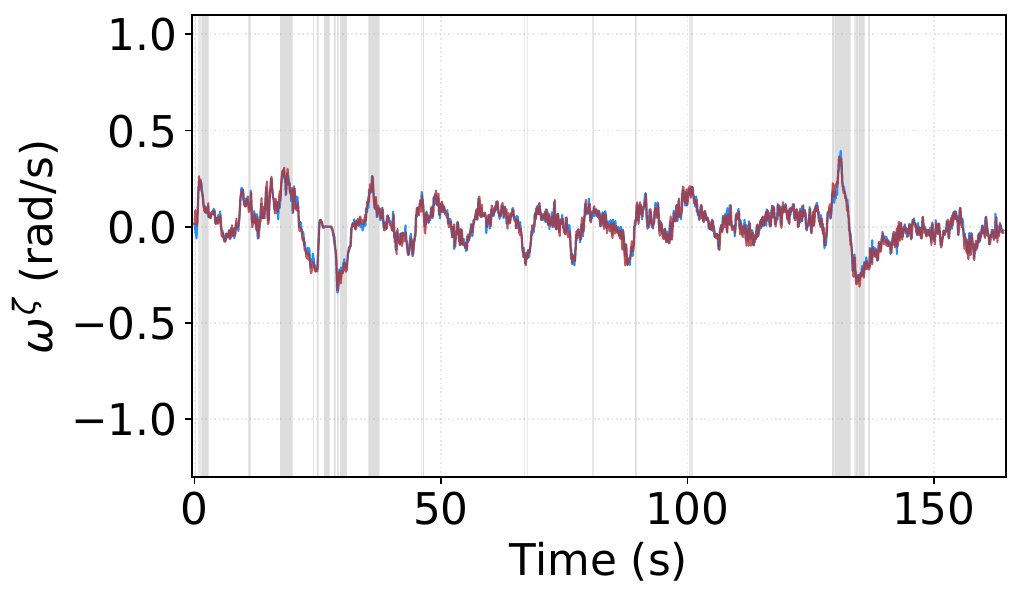} }
    \hfill \subfloat[Flight Case.\label{fig:nn_3_rw}]{ \includegraphics[width=0.21\textwidth]{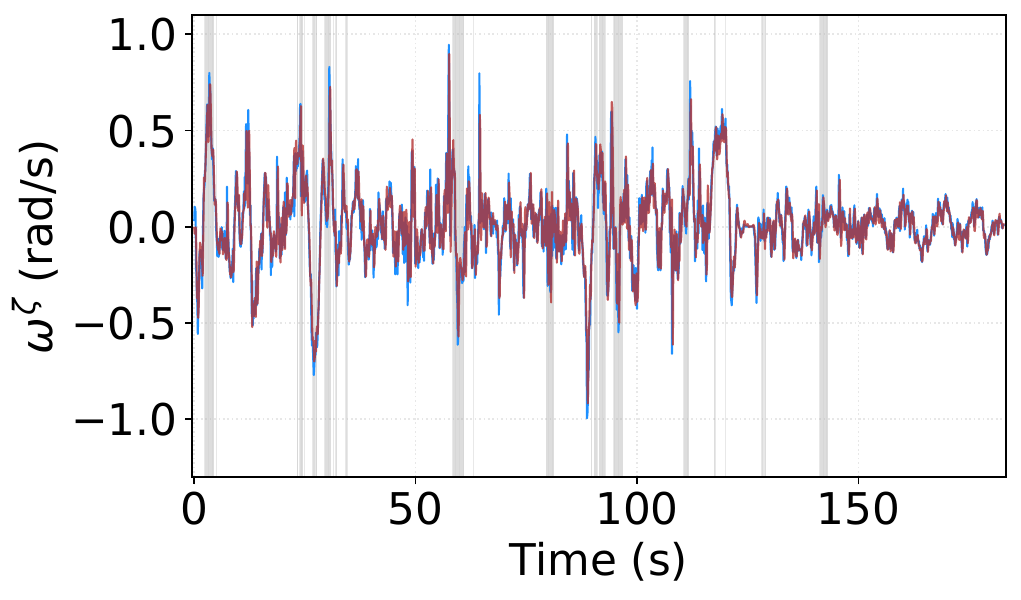} }
    \hfill \subfloat[Box.\label{fig:nn_4_rw}]{ \includegraphics[width=0.21\textwidth]{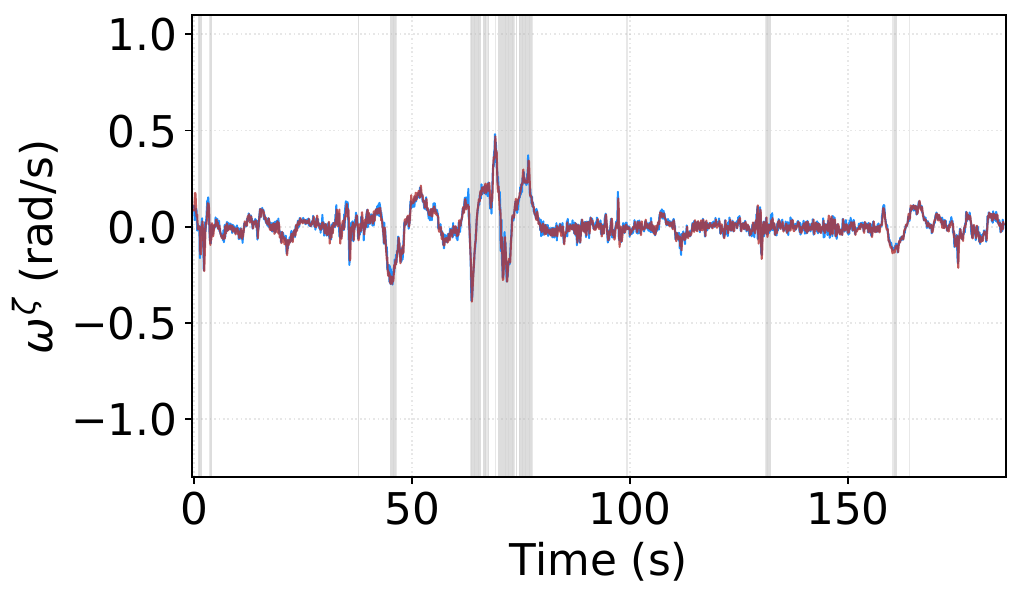} }
    \subfloat{ \raisebox{0.3\height}{\includegraphics[width=0.05\textwidth]{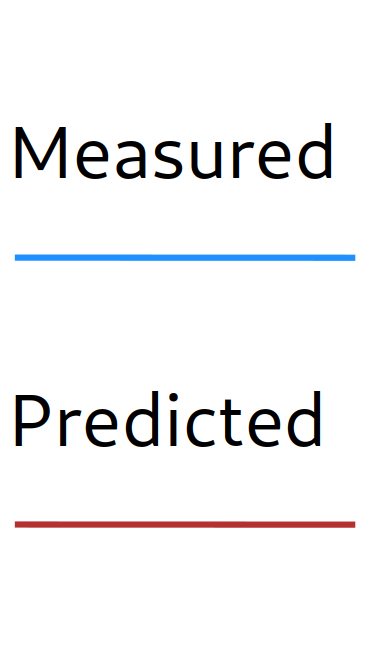}} }
    \vspace{-0.1cm}
    \caption{Prediction performance of the model in transportation
    experiments, with gray-shaded regions indicating activation of the residual kinematics
    ($s_{e}= 1$).}
    \vspace{-0.25cm}
    \label{fig:nn_rw}
\end{figure*}

\begin{figure*}[t]
    \centering
    \subfloat[Suitcase.\label{fig:ctrl_1_rw}]{ \includegraphics[width=0.235\textwidth]{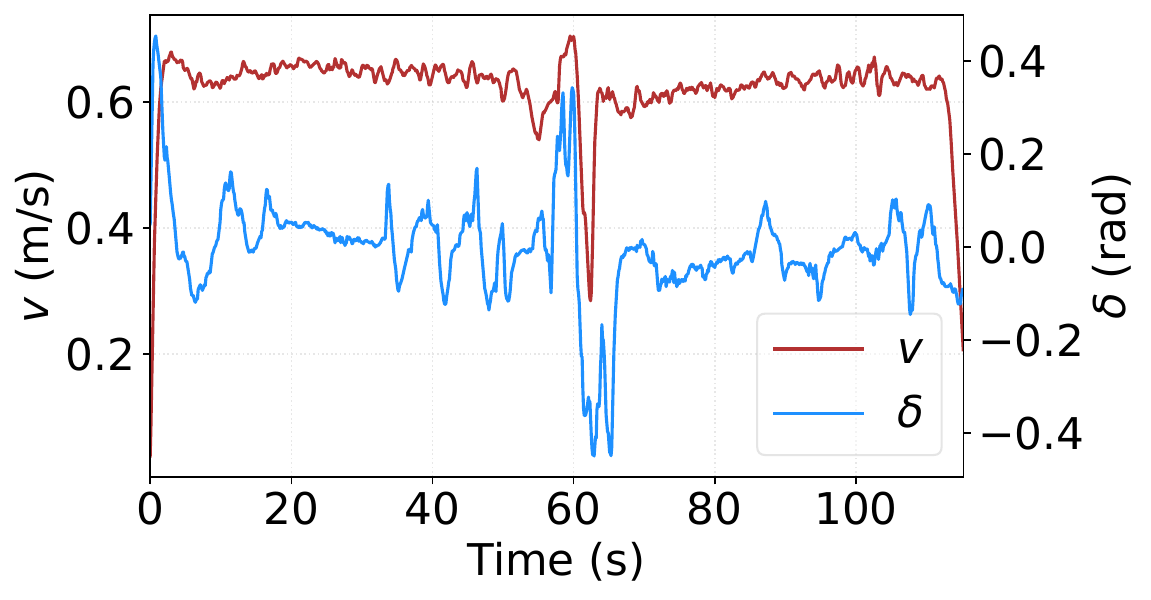} }
    \hfill \subfloat[Platform Trolley.\label{fig:ctrl_2_rw}]{ \includegraphics[width=0.235\textwidth]{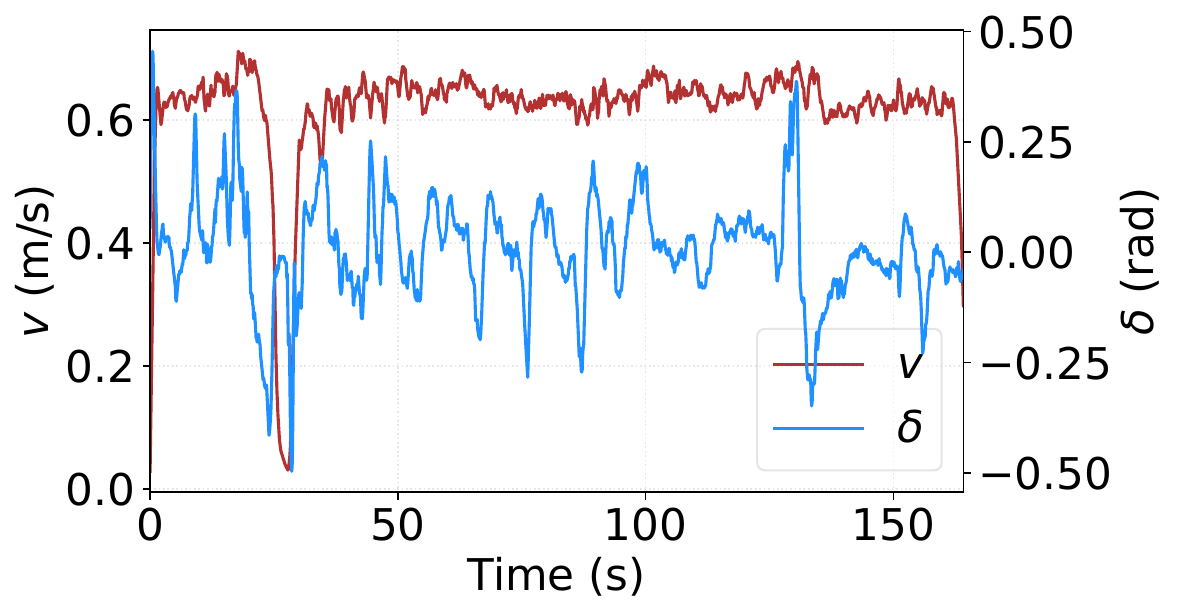} }
    \hfill \subfloat[Flight Case.\label{fig:ctrl_3_rw}]{ \includegraphics[width=0.235\textwidth]{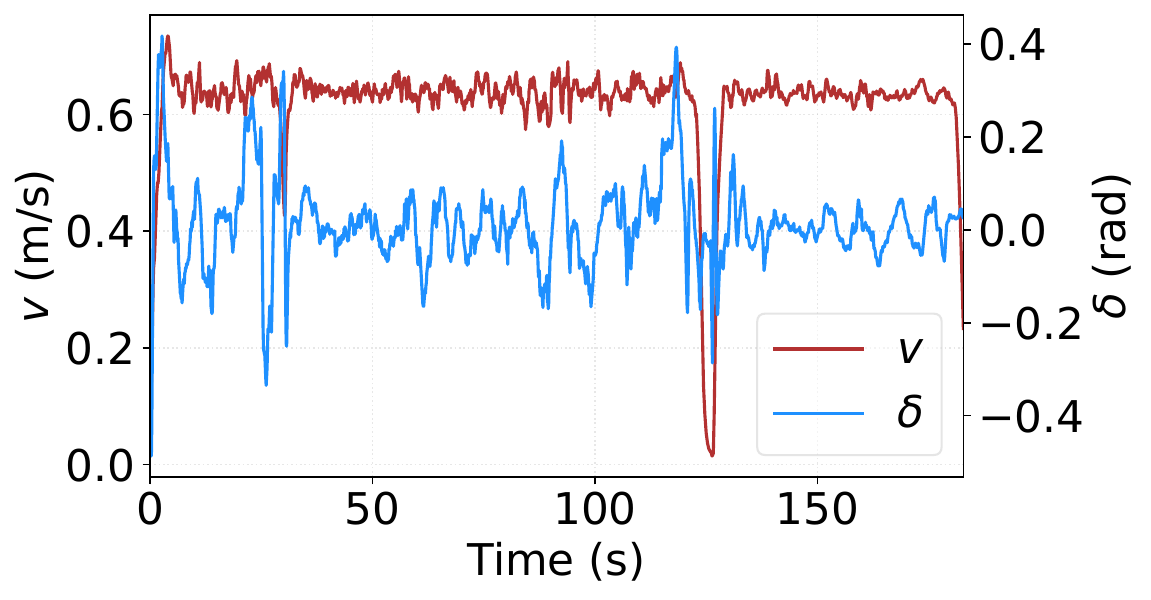} }
    \hfill \subfloat[Box.\label{fig:ctrl_4_rw}]{ \includegraphics[width=0.235\textwidth]{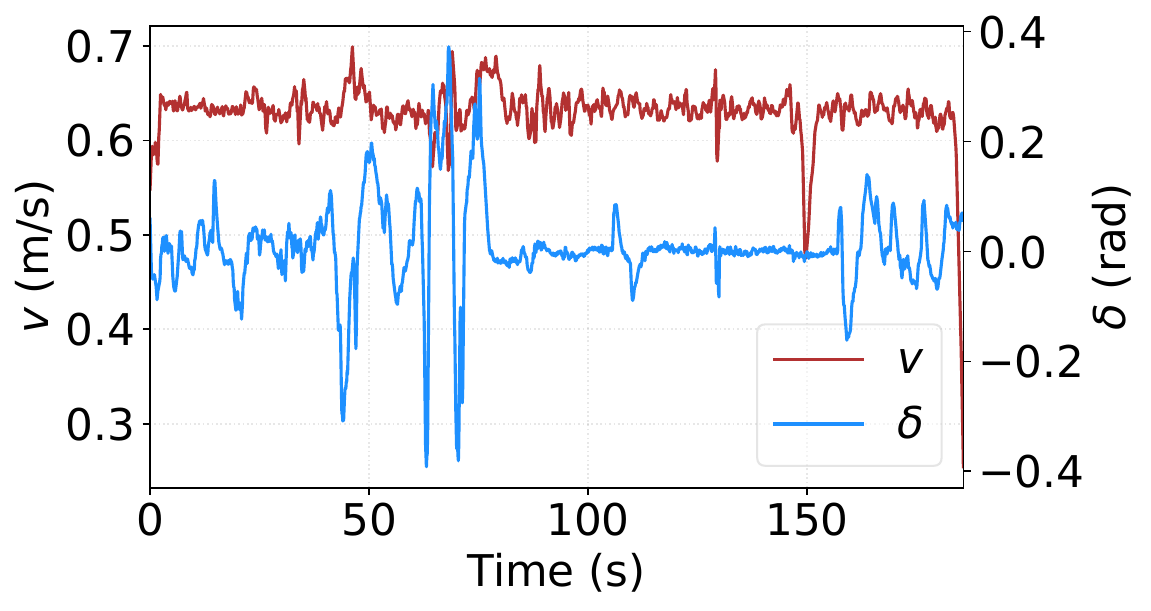} }
    \vspace{-0.1cm}
    \caption{Control inputs of the Ackermann-steering vehicle during trailer
    transportation experiments.}
    \label{fig:ctrl_rw}
    \vspace{-0.25cm}
\end{figure*}

Fig.~\ref{fig:nn_rw} and Fig.~\ref{fig:ctrl_rw} illustrate the neural network's prediction
performance and system control inputs during the experiments, respectively. The nominal
model effectively captures the system kinematics under most conditions and exhibits
good generalization capability for untrained trailers, such as the suitcase and
box, during straight-line motion. However, during turns, noticeable prediction
errors occur, triggering the activation of the residual kinematic model (e.g.,
between 0--8\,s and 58--68\,s in Fig.~\ref{fig:nn_1_rw}, and between 58--78\,s in
Fig.~\ref{fig:nn_4_rw}). For the flight case, the uneven terrain (See Fig.~\ref{fig:real_world_snapshot}(f),
(g), and the supplemental video) introduces high-frequency disturbances, and
they are effectively handled through multiple short-term activations of the
residual kinematic model (See Fig.~\ref{fig:nn_3_rw}), highlighting the system's
adaptability and robustness.

\section{Conclusion and Future Work}
\label{sec:conclusion}

This paper presents a novel universal vehicle-trailer navigation system, capable of autonomously adapting to varying types of trailers, payloads, and environmental disturbances. The key innovations include: a hybrid nominal model combining non-holonomic vehicle constraints with neural network-based trailer kinematics; an online residual learning module that corrects modeling errors in real-time; and an MPC framework with a weighted model combination strategy that enhances long-horizon prediction accuracy and safety. Extensive real-world experiments validate the robustness of the system, demonstrating accurate trajectory tracking and reliable obstacle avoidance across various trailer types.

Future work will focus on further automating the vehicle-trailer interaction process. 
Currently, changing between different trailers still requires manual intervention. 
We aim to design a dedicated mechanism and develop corresponding capture strategies to enable fully automated attachment and detachment of different trailers. Additionally, extending our system's capabilities to handle high-speed scenarios and dynamic obstacles constitutes another promising research direction. These enhancements strengthen the system's autonomy and practicality for real-world applications.


\newpage

\bibliographystyle{IEEEtran}
\bibliography{IEEEabrv,bib/main}
\end{document}